\pdfoutput=1

\documentclass[11pt]{article}

\usepackage{ACL2023}

\usepackage{times}
\usepackage{latexsym}
\usepackage{amsmath}
\usepackage{graphicx}
\usepackage[justification=centering]{caption}

\usepackage[T1]{fontenc}

\usepackage[utf8]{inputenc}

\usepackage{microtype}

\usepackage{inconsolata}

\title{Making the Most Out of the Limited Context Length: \\ Predictive Power Varies with Clinical Note Type and Note Section}

\author{Hongyi Zheng\textsuperscript{1} \quad Yixin Tracy Zhu\textsuperscript{1} \quad  Lavender Yao Jiang\textsuperscript{1, 2} \\
\bf{Kyunghyun Cho\textsuperscript{1}} \quad \bf{Eric Karl Oermann\textsuperscript{1, 2}} \\
    NYU Center for Data Science\textsuperscript{1} \quad NYU Langone Health\textsuperscript{2} \\
    \texttt{\{hz2212, yz5880, lyj2002, kyunghyun.cho\}@nyu.edu, \texttt{eric.oermann@nyulangone.org}}
}

\begin{document}
\maketitle
\begin{abstract}
    Recent advances in large language models have led to renewed interest in natural language processing in healthcare using the free text of clinical notes. One distinguishing characteristic of clinical notes is their long time span over multiple long documents. The unique structure of clinical notes creates a new design choice: when the context length for a language model predictor is limited, which part of clinical notes should we choose as the input? Existing studies either choose the inputs with domain knowledge or simply truncate them. We propose a framework to analyze the sections with high predictive power. Using MIMIC-III, we show that: 1) predictive power distribution is different between nursing notes and discharge notes and 2) combining different types of notes could improve performance when the context length is large. Our findings suggest that a carefully selected sampling function could enable more efficient information extraction from clinical notes. 
\end{abstract}

\section{Introduction}

Electronic Health Records (EHR) enable the development of language model based clinical predictor, which takes in clinical notes to predict patient outcomes. Clinical notes in EHR exhibit two unique characteristics. 1) Clinical notes cover a long time span (from a few weeks to over a year), which results in their sparsity of information-rich sections. 2) Clinical notes also tend to be long: many discharge notes could take up to $10,000$ tokens, which makes using the entire note as model input computationally expensive. 3) The strong noise level in the medical notes (usually due to the domain-specific abbreviations and typos) also poses a challenge to extract information effectively.

These distinguishing characteristics of clinical notes lead to a new design choice: when the context length is limited due to the constrained compute or model architecture, what parts of clinical notes should we sample to maximize the model's performance? We propose a framework to subsample text sections with high predictive power. 

Empirically, we explore the distribution of predictive power over clinical note types and sections by searching over these variables. We found that 1) the predictive power distribution is different between nursing notes and discharge notes: the predictive power is stronger at the beginning and end of discharge notes, while uniform within nursing notes. 2) The effect of combining sections from different types of notes improves the performance when the context size is large, but harms the performance when the context size is small. More details of task formulation can be found at \autoref{method}. Our code is publicly available on GitHub\footnote{\url{https://github.com/nyuolab/EfficientTransformer}}.


\section{Related Work}


Existing methods for subsampling clinical notes for the BERT-based model are mostly based on domain knowledge. For instance, \citet{yang2022language} and \citet{9056492} choose discharge notes as they summarize patients' visits.  \citet{thapa2022hospital} chooses the notes within three days before a cutoff time in consideration of timeliness. While these assumptions are based on domain knowledge, they require human input and may not generalize. Thus, we are interested in exploring a data-driven sampling choice without assumptions of expert inputs.

Another related, but orthogonal approach to the limited context length problem is note aggregation. Instead of subsampling notes, \citet{huang2019clinicalbert} propose to feed everything to the model, one maximum context length at a time, and aggregate the outputs for the final prediction. In their work, notes of one patient are split into a partition of subsequences, and the patient's re-admission risk is obtained by taking a weighted average of probabilities computed from each subsequence. This method's compute cost scales with the aggregated sequence length, which can be expensive for records with long clinical notes. In contrast, our method aims to find one single information-rich segment as input.



\section{Method}\label{method}

We formalize our prediction task as follows: given a set of clinical notes $x$ associated with an admission record, we want to predict the class label $y$ which is our patient outcome of interest. Ideally, we want to train a classifier $f_{w^*}$ to approximate $p(y \mid x)$. The optimal parameter is
\begin{equation*}
    w^* = \arg\max_w ~ m (f_w(x), y),
\end{equation*}
where $m$ is a metric function of interest. Nevertheless, due to the computational constraint, we need to reduce the input size via a sampling function $s_\theta$ so that $s_\theta(x)$ fits the input length limit and preserves information. Empirically, the optimal parameters are
\begin{equation*}
    w^*, \theta^* = \arg\max_{w, \theta} ~ m(f_w(s_\theta(x), y)).
\end{equation*}

We say a sample function $s_\theta$ has a higher predictive power if $m(f_{w^*}(s_\theta(x), y))$ is larger. 

While current works chose $s_\theta$ based on prior medical knowledge or simply fix it as a truncation function, we propose to explore different sampling functions $s_\theta$ to make the most out of the limited context length with the highest predictive power. Notice that in our work, $s$ and $\theta$ are searched manually, instead of using learning algorithms.


\section{Experimental Setup}

We hypothesize that for 30-day all-cause readmission prediction, there exists an alternative sampling function that enables similar or better performance than the commonly used ``truncated discharge notes". More formally, we focus on a parameterized sampling function with 2 variables: 1) which section of tokens to include, 2) what type(s) of clinical notes to use.

\paragraph{Model} We finetuned two clinical language models in our experiments. The first is Clinical-BERT \citep{alsentzer2019publicly}, which continued to pretrain BERT using approximately 2 million notes from MIMIC-III and has a maximum sequence length of 512. The second is the ClinicalLongformer \citep{DBLP:journals/corr/abs-2201-11838}, which continued to pretrain Longformer \citep{DBLP:journals/corr/abs-2004-05150} with MIMIC-III notes and enables input of up to $4096$ tokens. Both models are finetuned to predict the probability of 30-day all-cause readmission: that is, whether the patient will be re-admitted to the hospital within 30 days of their discharge dates. 

\paragraph{Dataset} We use the discharge notes and nursing notes in the \texttt{noteevent} table of the MIMIC-III database \cite{mimic}. There are $~40,000$ de-identified admission records available to use after filtering out all admission records without nursing notes and discharge notes. The admission records are split into 75\% train, 12.5\% validation, and 12.5\% test sets. Other types of medical notes such as physician notes are excluded from consideration in our experiments due to their scarcity in the database. See Appendix \ref{app:processing} for data preprocessing.

\paragraph{Sliding Window} To extract different sections of the clinical notes, we use a sliding window technique. Let $n$ be the window's width. Let $l$ be the total number of tokens of the text. The window is placed based on an input parameter $p \in [0, 1]$ indicating the location of the midpoint of the window, where the window interval is 
\begin{equation*}
    [lp - n / 2, lp + n / 2].
\end{equation*}
In case where $lp - n / 2 < 0$, we shifted the window backward so that the front of the window aligns with the beginning of the input tokens. In the case where $lp + n / 2 > l$ we shifted the window forward to let the back of the window match the end of the tokens. Also, when $l < n$, we ignore the input $p$ and pad the tokens to maximum input length $n$. 

We try $11$ different values of $p$ ($0.0, 0.1, \cdots 1.0$) for ClinicalBERT and $2$ values of $p$ ($0.0$ and $1.0$) for ClincialLongformer along with an additional fragmented window trial $p = \mathrm{both}$ which looks into the first $n / 2$ and last $n / 2$ tokens of the input text. Similarly, when $l < n$, we simply pad the sequence to the window's length.

\paragraph{Mixing Notes}  To control different types of clinical notes, we experimented with the following options: 1) first nursing note, 2) last nursing note, 3) discharge note, 4) first nursing notes + discharge note, 5) last nursing notes + discharge notes. For options with two types of notes, $n / 2$ tokens are allocated to each type, and three values for $p_1$ and $p_2$ each ($0.0, 1.0$ and $\mathrm{both}$) are used to select $n / 2$ tokens from each type of note, resulting in $9$ possible input parameter combinations. 

\begin{figure*}[t]
    \centering
    \includegraphics[width=.9\textwidth]{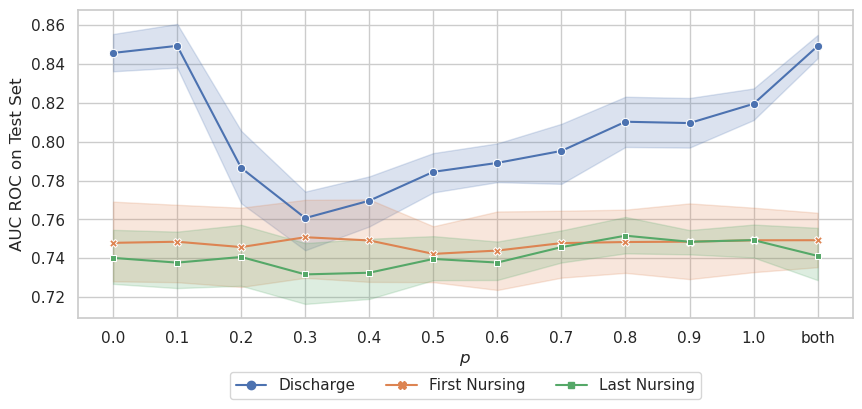}
    \caption{
    Performance of ClinicalBERT on Different Text Sections and Different Types of Notes, \\
    Error Bars Represent 95\% Confidence Intervals
    }
    \label{fig:clinicalbert}
\end{figure*}

\section{Results}

\subsection{Different Sections in Nursing Notes and Discharge Notes}\label{sec:nursing_vs_dc}

We finetune ClinicalBERT and ClinicalLongformer on different sections of nursing and discharge notes. We used sliding windows to extract a sequence of tokens that meets the model's maximum sequence length. We have three key observations.\\

\noindent\textbf{Different Types of Clinical Notes Show Disparate Predictive Power Distributions Over Text Sections}. As shown in \autoref{fig:clinicalbert}, the discharge notes (blue line) show quite uneven predictive power distribution, where the beginning ($p=0.0$) and end ($p=1.0)$ sections of the text provide strong predictive power while the middle sector ($0.2\leq p \leq 0.5$) shows a significant dip in predictive power. In contrast, the predictive power of the nursing notes (orange and green line) turns out to be uniformly distributed: using different sections of the nursing notes ($0.0\leq p\leq 1.0$) does not make a significant difference. We speculate that this discrepancy may stem from the domain knowledge that discharge notes are more structured than nursing notes: they often start with basic descriptions of the patient information and ends with suggestions for the patients, whereas nursing notes often have multiple types of information mixed together throughout the text.\\

\begin{figure}[h]
    \centering
    \includegraphics[width=0.48\textwidth]{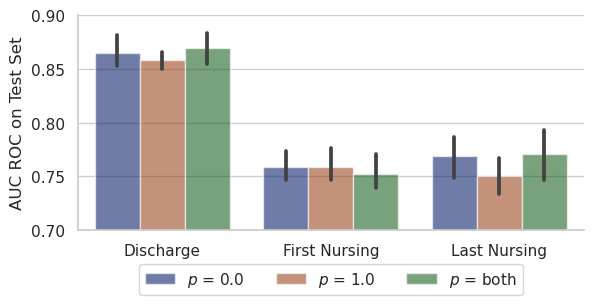}
    \caption{Performance of ClinicalLongformer on Different Text Sections and Different Types of Notes, \\
    Error Bars Represent 95\% Confidence Intervals}
    \label{fig:clinicallongformer}
\end{figure}

\noindent\textbf{Nursing Notes Provide Modest Predictive Power}. Nursing notes produce decent re-admission prediction results: according to \autoref{fig:clinicalbert} and \autoref{fig:clinicallongformer}, although their predictive power is not as strong as discharge notes (which are typically written right before patients leave the hospital), they consistently achieve AUC ROC scores of over $0.7$ which indicates modest predictability \cite{schneeweiss2001performance}. Moreover, the first nursing notes (orange line in \autoref{fig:clinicalbert}, second group of bars in \autoref{fig:clinicallongformer}) of each admission provide similar predictive power as compared to the last nursing notes (green line in \autoref{fig:clinicalbert}, third group of bars in \autoref{fig:clinicallongformer}), indicating the possibility of re-admission risk evaluation at the early stage of the admission. This finding is especially valuable from the perspective of intervention, as it is more practical to decide whether the patient should be discharged at the time before the discharge note is written. Also, the abundance of nursing notes makes them a suitable alternative for re-admission risk evaluation tasks when discharge notes are unavailable.

\begin{figure*}[t]
    \centering
    \includegraphics[width=.9\textwidth]{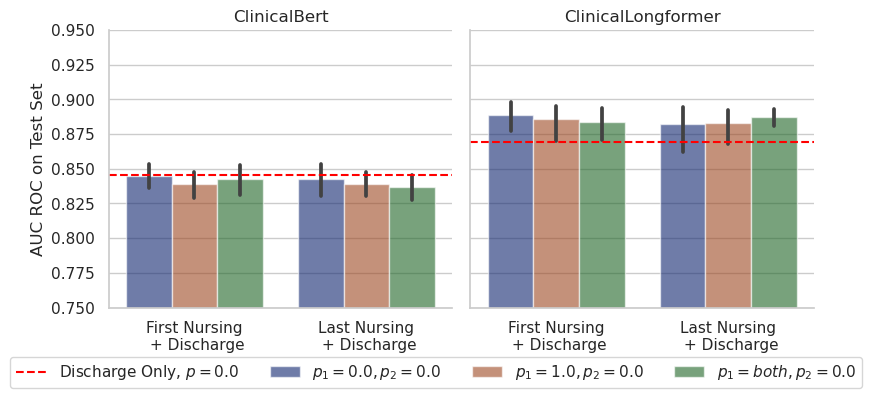}
    \caption{Performance of ClinicalBERT and ClinicalLongformer on Clinical Note Combinations, \\
    Error Bars Represent 95\% Confidence Intervals}
    \label{fig:clinicallongformermix}
\end{figure*}

\noindent\textbf{Preserving the Beginning Tokens Is Not the Only Option}. It is generally assumed that when the available input tokens are limited, the leading tokens of each clinical note should be used. Nevertheless, our experiments show that for discharge notes, spending half of the available tokens on the beginning section and spending the remaining half on the end section ($p = \mathrm{both}$) achieves slightly better performance (AUC ROC of $0.849$ versus $0.845$ for ClinicalBERT, $0.869$ versus $0.864$ for ClinicalLongformer) as compared to using the leading token only ($p = 0.0$). We speculate that this helps as it avoids the weakly predictive middle sector of the clinical notes.


\subsection{Combining Sections from Different Types}\label{sec:combine_notes}

We combine text sections from two different types of clinical notes and finetune ClinicalBERT and ClinicalLongformer. This experiment helps us investigate the question: when the amount of available tokens is fixed, does combining information from different clinical notes work better than using discharge notes only? Since discharge notes are shown to provide strong predictive power in our prior experiments, we only investigate the note type combinations that include discharge notes (first nursing + discharge, last nursing + discharge). \\


\noindent\textbf{The Effect of Allocating Tokens to Different Types of Clinical Notes Depends on the Context Size}. When the context size is relatively large (ClinicalLongformer, as shown in the right side of figure \ref{fig:clinicallongformermix}), allocating the available tokens to different types of clinical notes (blue, orange, and green bars) leads to improvements in performance. The baseline (dashed red line) uses discharge notes only and has a lower AUC ROC ($0.013$ to $0.019$) than models finetuned with combined notes. However, when the context is small (Clinical BERT, as shown in the left side of figure \ref{fig:clinicallongformermix}), distributing the already limited number of tokens to different clinical notes hurts the performance: the AUC ROC of ClinicalBERT finetuned with mixed notes falls below the baseline performance by $-0.009$ to $-0.001$. We speculate that this may be related to the uneven predictive power distribution in discharge notes: if there are already a sufficient number of tokens covering the most informative sections of the discharge notes, the rest of the discharge notes might not be as informative as the prior nursing notes. 

\section{Discussion and Future Works}

Our findings suggest that when the input size is constrained, a carefully selected sampling function that chooses the text with high predictive power could benefit model performance. Specifically on the task of readmission prediction from MIMIC-III notes, we show that the predictive power varies across note types and note sections. This insight enables more efficient information extraction from long and noisy clinical notes, which is beneficial when the computing resource is limited and the context length needs to be controlled.


Our findings call for two future directions. First, the performance disparities between ClinicalBERT and ClinicalLongformer (\autoref{sec:combine_notes}) indicate that the best strategy to allocate the input context is related to the maximum sequence length, and more work should be done to determine their exact relationship. Another direction is investigating the predictive power pattern based on the authorship of the clinical note. We showed (\autoref{sec:nursing_vs_dc}) that discharge notes (written by doctors) have a more uneven predictive power pattern as compared to nursing notes (written by nurses). How the domain knowledge of the author would affect the clinical note quality is worth investigating.

\section*{Limitations}

We acknowledge three limitations in our experiments. First, in our second experiment, we fixed the window size for each type of note to be $n / 2$. A more comprehensive investigation could also search for the optimal window size for each note type. Second, although we explored one fragmented window configuration $p = \mathrm{both}$, we did not explore other fragmented window configurations due to resource constraints. Lastly, we did not investigate more types of clinical notes (e.g., physician notes and ECG notes) because MIMIC-III has limited examples for other note types. We expect it to be resolved in future works with MIMIC-IV's publication \cite{mimiciv}.

\bibliography{anthology,custom}
\bibliographystyle{acl_natbib}

\section*{Appendices}
\appendix

\section{Preprocessing}\label{app:processing}

We preprocessed the dataset with the following approach: First of all, admission records with missing discharge notes or missing nursing notes are eliminated. Then, for each remaining admission record, the nursing notes associated with that record are sorted according to their timestamp. The first and last created nursing notes for each admission are selected and concatenated with the discharge notes of the same admission record to produce the clinical note set for every admission. Lastly, we clean the datasets by removing the de-identification patterns ('[** de-identified info **]') in the clinical notes, which usually occupy a lot of tokens.

\end{document}